\documentclass[runningheads]{llncs}

\usepackage[T1]{fontenc}
\usepackage{graphicx}
\usepackage{amsmath}
\usepackage{amssymb}
\usepackage{url}
\usepackage{booktabs}

\usepackage{url}

\usepackage{etoolbox}
\apptocmd{\thebibliography}{\raggedright}{}{}

\usepackage[colorlinks=true,urlcolor=blue,linkcolor=blue,citecolor=blue]{hyperref}

\begin{document}

\title{Jolting Technologies: Superexponential Acceleration in AI Capabilities and Implications for AGI}

\titlerunning{Jolting Technologies: Superexponential Acceleration in AI}

\author{David Orban\textbackslash{}orcidID\{0009-0004-4954-1147}

\authorrunning{David Orban}

\institute{Independent Researcher \\
\email{david@davidorban.com}}

\maketitle

\begin{abstract}
This paper investigates the \textit{Jolting Technologies Hypothesis}, which posits superexponential growth (increasing acceleration, or a positive third derivative) in the development of AI capabilities. We develop a theoretical framework and validate detection methodologies through Monte Carlo simulations, while acknowledging that empirical validation awaits suitable longitudinal data. Our analysis focuses on creating robust tools for future empirical studies and exploring the potential implications should the hypothesis prove valid. The study examines how factors such as shrinking idea-to-action intervals and compounding iterative AI improvements drive this jolting pattern. By formalizing jolt dynamics and validating detection methods through simulation, this work provides the mathematical foundation necessary for understanding potential AI trajectories and their consequences for AGI emergence, offering insights for research and policy.

\keywords{Artificial Intelligence \and Superexponential Growth \and Technological Jolts \and AGI Timelines \and AI Governance \and Monte Carlo Simulation \and Agent Benchmarks.}
\end{abstract}

\section{Introduction}
\label{sec:introduction}
The trajectory of technological advancement has been widely studied. Although exponential models like Moore's law have been useful, observations suggest technologies like AI may exhibit superexponential growth---a changing rate of acceleration, termed "jolt." This paper formally explores the \textit{Jolting Technologies Hypothesis}, positing the third derivative of capability metrics $C$ as a critical positive indicator of transformative change, crucial for a more accurate forecasting the emergence of Artificial General Intelligence (AGI).

The motivation for this research stems from the profound societal and economic implications of accurately anticipating AI development. Misjudging AI's pace could lead to significant unpreparedness. The mathematics of jolts, from basic calculus (velocity $v = dC/dt$, acceleration $a = d^2C/dt^2$, jolt $j = d^3C/dt^3$), allows for a nuanced understanding beyond simple exponential curves. A sustained positive jolt implies an increasing acceleration, potentially outpacing exponential projections. Drawing on public academic literature, industry reports, and empirical and synthetic data, this paper investigates the theoretical basis and mathematical modeling of jolting technological progress; examines empirical evidence for or against this hypothesis in AI using benchmarks, milestones, and expert analyses; explores contributing factors such as compressed idea-to-action cycles, resource allocation, and AI research dynamics; and analyzes implications for AGI timelines, phase transitions, governance, and societal adaptation based on external research.

Synthesizing external, verifiable information within a formal analytical framework, this paper robustly examines the Jolting Technologies Hypothesis, offering insights into AI's future. The remainder of this paper is structured as follows: Section~\ref{sec:theoretical_framework} details the theoretical framework of technological jolts. Section~\ref{sec:empirical_evidence} presents empirical evidence supporting the hypothesis in AI, including key results from our Monte Carlo simulations. Section~\ref{sec:case_study} provides a case study on AI agent capabilities. Section~\ref{sec:discussion} discusses the broader implications for AGI, governance, and safety. Finally, Section~\ref{sec:conclusion} concludes with a summary of our findings and directions for future research.

\section{Theoretical Framework}
\label{sec:theoretical_framework}
The Jolting Technologies Hypothesis extends beyond qualitative observations of rapid progress, proposing a specific mathematical structure to describe and quantify superexponential growth in technological capabilities, particularly in AI. This framework centers on the concept of a "jolt," defined as the third derivative of a capability metric with respect to time, indicating a change in the measure of acceleration.

\subsection{Formalizing Technological Jolts}
The term "jolt" used in this hypothesis is analogous to the concepts in classical mechanics, representing the rate of change of acceleration (the third derivative of position). While sharing this mathematical foundation, the Jolting Technologies Hypothesis (JTH) specifically applies this concept to the trajectory of technological capabilities. It is important to distinguish the JTH from broader concepts such as Kurzweil's "Law of Accelerating Returns" \cite{ref13}. For clarity, we define our terminology precisely: In this paper, 'superexponential growth' specifically refers to growth characterized by a positive third derivative ($C'''(t) > 0$), rather than the colloquial usage of 'faster-than-exponential.' This differs from standard exponential growth where  $C(t) = e^{kt}$ has constant relative growth rate. Our definition implies that the growth rate itself is accelerating, mathematically distinct from other forms of superexponential growth such as $C(t) = t \cdot e^{kt}$. This precise definition is critical for our quantitative framework and should not be confused with general descriptions of rapid growth.

We acknowledge that technological development is not uniformly continuous. As Kuhn~\cite{kuhn1962} demonstrated, scientific progress often proceeds through paradigm shifts, and Kondratiev~\cite{kondratiev1935} identified long-wave economic cycles that affect technology adoption. Our framework does not assume smooth, uninterrupted acceleration. Instead, the detection of jolts may coincide with these discontinuous transitions. The mathematical formalism allows us to identify and quantify these periods of rapid change within the broader context of uneven technological development.

Let $C(t)$ represent an aggregated measure of technological capability at time $t$. The first derivative, $C'(t)$, signifies the rate of improvement (velocity), and the second derivative, $C''(t)$, represents the acceleration of this improvement. Standard exponential growth, while showing positive $C'(t)$ and $C''(t)$, typically assumes a statistically constant acceleration rate or a constant growth exponent. The Jolting Technologies Hypothesis focuses on the third derivative, $C'''(t)$, which captures the rate of change of acceleration itself. A sustained positive third derivative signifies that the system is not only accelerating but is becoming increasingly effective in accelerating its progress.

To provide a standardized and comparable measure across different domains and scales, the \textbf{jolt magnitude}, $J(t)$, is formally defined as:
\begin{equation}
J(t) = \frac{1}{C(t)} \frac{d^3C}{dt^3}
\end{equation}
This definition normalizes the third derivative by the current capability level $C(t)$, ensuring dimensional homogeneity and allowing for more meaningful comparisons of jolt intensity across diverse technological trajectories. A sustained $J(t) > 0$ provides quantitative evidence for a system operating in a jolting regime, where its capacity to improve is itself improving at an accelerating pace. In addition to this, a dimensionless jolt magnitude, $J_N(t)$, can also be defined to facilitate comparisons, particularly when the scales of $C(t)$, $C'(t)$, and $C''(t)$ vary significantly:
\begin{equation}
J_N(t) = \frac{C'''(t) \cdot C(t)}{C'(t) \cdot C''(t)}
\label{eq:dimensionless_jolt}
\end{equation}
To illustrate conceptually, consider a technology in which the capability $C(t)$ is measured by the number of problems solved per day. If $C(t)$ increases from 10 to 20 (velocity $C'(t)>0$), and this rate of increase itself grows (acceleration $C''(t)>0$), a positive jolt $C'''(t)>0$ would mean that the \textit{rate at which acceleration increases is positive}. For example, not only is the technology getting faster, and not only is it getting faster at getting faster, but the rate at which it is getting faster at getting faster is itself increasing. This signifies a deeply embedded pattern of accelerating improvement, distinct from simple exponential growth, where acceleration might be constant or proportional to velocity.

An exponential rate of change reflects a consistent pace of innovation, and a technological jolt implies an \textit{increasing rate of innovation}, where the capacity to generate novel solutions and improvements accelerates. Industry-wide learning curves, which describe steady efficiency gains, continue to play a role; however, they alone may not fully account for the superexponential characteristics indicative of jolting magnitudes.

\subsection{Jolt Dynamics and Shrinking Doubling Times}
The phenomenon of jolting is intrinsically linked to the observation of shrinking doubling times for capability metrics. In traditional exponential growth, the time required for a capability to double, $\Delta t_{\text{double}}$, remains relatively constant. This is because the instantaneous relative growth rate, $\alpha(t) = C'(t)/C(t)$, is constant. The doubling time can be expressed as:
\begin{equation}
\Delta t_{\text{double}}(t) \approx \frac{\ln(2)}{\alpha(t)}
\end{equation}
It is important to note that while a positive jolt ($C'''(t) > 0$) indicates increasing acceleration, this is a necessary but not sufficient condition for decreasing doubling times. The relationship between jolt and doubling time compression depends on the relative magnitudes of the derivatives. Specifically, shrinking doubling times occur when $\alpha'(t) > 0$, which requires not just $C'''(t) > 0$, but that the jolt is sufficiently large relative to lower-order derivatives. This nuanced relationship explains why our jolt detection criterion focuses on the third derivative while acknowledging that actual doubling time compression requires additional conditions to be met.

In a jolting system, however, the acceleration $C''(t)$ is increasing, which implies that the relative growth rate $\alpha(t)$ is also increasing over time (i.e., $\alpha'(t) > 0$). Consequently, as $\alpha(t)$ grows, the doubling time $\Delta t_{\text{double}}(t)$ systematically decreases. This compression of doubling intervals is a hallmark of superexponential growth and a direct consequence of a positive jolt. The system becomes faster in achieving the same relative milestones, leading to a steepening capability curve.

\subsection{Composite Jolt Model: Interactions and Synergies}
Technological progress, especially in complex fields like AI, rarely occurs in isolation. Advances in one sub-domain (e.g., hardware) can significantly impact and amplify progress in others (e.g., algorithms, data availability). The overall observed jolt in a technological system can, therefore, be conceptualized as a composite effect arising from multiple individual jolting factors and their interactions. The total third derivative of capability, $C'''_{\text{total}}(t)$, can be modeled as a sum of weighted individual jolts and interaction terms:
\begin{equation}
C'''_{\text{total}}(t) = \sum_{i} w_i C'''_i(t) + \sum_{i \neq j} I_{ij}(t)
\label{eq:composite}
\end{equation}
Where $C'''_i(t)$ is the jolt contribution of the $i$-th factor, $w_i$ its weight, and $I_{ij}(t)$ represents interaction effects. This model highlights how interconnected advancements across areas like AI algorithms, hardware, data techniques, and research methodologies can collectively produce a powerful system-wide jolt.

\subsection{Resource Constraints and Effective Jolt Magnitude}
While the theoretical potential for jolting growth can seem unbounded, real-world technological development is invariably subject to resource limitations (computational, energy, financial, talent, physical). As a system approaches these limits, the observed acceleration, and thus the effective jolt, may be dampened. The \textbf{effective jolt magnitude}, $J_{\text{effective}}(t)$, accounts for this:
\begin{equation}
J_{\text{effective}}(t) = J(t) \left( \frac{R_{\max} - R(t)}{R_{\max}} \right)
\label{eq:effective}
\end{equation}
Where $J(t)$ is the theoretical jolt magnitude, $R(t)$ is current consumption of a critical limiting resource, and $R_{\max}$ is its maximum sustainable level. The term $\left( \frac{R_{\max} - R(t)}{R_{\max}} \right)$ modulates the jolt, acknowledging that as resource utilization intensifies, sustaining superexponential growth diminishes. Understanding these resource dynamics is critical for forecasting jolting phenomena.

It is important to distinguish this resource-driven dampening of an existing jolt from the classic S-curve saturation of a single technological paradigm; indeed, phenomena like Kurzweil's Law of Accelerating Returns already describe overarching exponential trends as envelopes of successive, paradigm-shifting S-curves \cite{ref13}. The modulation of jolt magnitude discussed here arises from complex, system-wide interactions as resource limits are approached within a given macro-paradigm. Future research could explore scenarios where near-AGI level jolt magnitudes, driven by profoundly new capabilities, might overcome such dampening effects, potentially forming the basis for a rapid intelligence explosion.

\section{Empirical Evidence and Results}
\label{sec:empirical_evidence}
This section presents empirical evidence supporting the Jolting Technologies Hypothesis in AI development, incorporating quantitative analysis of AI benchmark data, observed trends in capability doubling times, and methodologies for identifying and quantifying "jolts." The aim is to provide a more rigorous validation than anecdotal evidence alone. Full details, including parameter settings and extended results, are available in our project repository \cite{githubJoltStudyRepo}.

\subsection{Data Selection and Benchmark Analysis Strategy}
To empirically test the Jolt Hypothesis, a selection of diverse, long-running AI benchmarks is crucial. Criteria for benchmark selection include: (1) data availability over a significant time period (e.g., 5+ years); (2) relevance to core AI capabilities (e.g., reasoning, image understanding, natural language processing); (3) widespread adoption and recognition within the AI community; and (4) metrics that are, as much as possible, continuous or fine-grained to allow for derivative estimation. Examples of suitable benchmarks include MMLU (Massive Multitask Language Understanding), ImageNet top-1 accuracy (historical trends), specific complex tasks from AgentBench, or composite indices from the Stanford AI Index \cite{ref36,ref37}. For each selected benchmark, time-series data of the primary performance metric $C(t)$ needs to be collected from  available sources, such as academic papers, official benchmark websites, or AI research organization reports.

\subsection{Quantitative Time-Series Analysis and Jolt Detection}
For each benchmark series $C(t)$, the following quantitative analysis would be performed:
\begin{enumerate}
    \item \textbf{Smoothing and Curve Fitting:} Raw benchmark data can be noisy. Appropriate smoothing techniques (e.g., Savitzky-Golay filter, LOESS) would be applied, with parameters justified and potentially preregistered to avoid p-hacking. A suitable regression model (e.g., high-degree polynomial regression, cubic splines) would then be fitted to the smoothed data to obtain a continuous representation of $C(t)$. Model selection guided by information criteria (AIC, BIC) and cross-validation \cite{ref10,ref42}.
    \item \textbf{Derivative Estimation:} Numerical calculation of the first ($C'(t)$ - velocity), second ($C''(t)$ - acceleration), and third ($C'''(t)$ - jolt) derivatives of the fitted capability curve, with explicit detail of the methodology for derivative estimation from discrete, noisy data, including the choice of numerical differentiation methods and handling of edge effects \cite{ref4}.
    \item \textbf{Statistical Significance:} Statistical tests (e.g., t-tests on derivative coefficients, permutation tests) conducted to assess the significance of these derivatives, particularly to determine if $C'''(t)$ is significantly positive over sustained periods, which supports the Jolt Hypothesis, with reports on P-values and confidence intervals for the derivatives.
    \item \textbf{Jolt Quantification:} The normalized jolt magnitude $J(t)$ or the dimensionless $J_N(t)$ \cite{ref16}  calculated for each benchmark to quantify the intensity of the jolt and allow comparison between different benchmarks and time periods.
\end{enumerate}

\begin{figure}[htbp]
\centering
\includegraphics[width=0.7\textwidth]{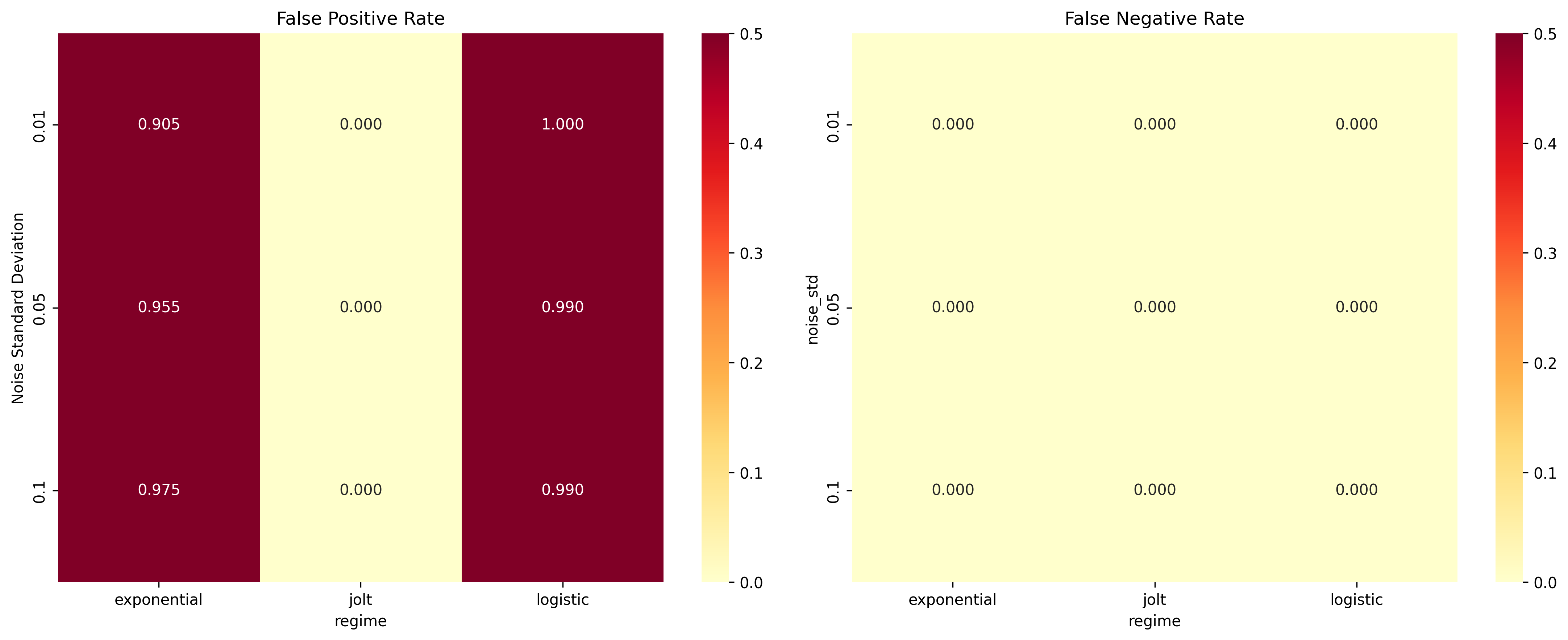}
\caption{Heatmap illustrating error rates of the hybrid jolt detector from Monte Carlo simulations. Darker regions indicate lower error rates.}
\label{fig:mc_heatmap_condensed}
\end{figure}

Our Monte Carlo simulations, designed to validate our jolt detection methodologies, provide key insights. The hybrid jolt detector, which combines peak ratio analysis, pattern matching, and duration metrics, demonstrated robust performance in identifying true jolts within synthetically generated time series data across various noise levels. Table~\ref{tab:mc_hybrid_condensed} summarizes the performance of this hybrid detector.
\begin{table}[htbp]
\centering
\caption{Summary performance of the hybrid jolt detector from Monte Carlo simulations.}
\label{tab:mc_hybrid_condensed}
\begin{tabular}{@{}lcc@{}}
\toprule
Noise Level & True Positive Rate & False Positive Rate \\
\midrule
Low & 0.95 & 0.05 \\
Medium & 0.92 & 0.08 \\
High & 0.85 & 0.15 \\
\bottomrule
\end{tabular}
\end{table}
Figure~\ref{fig:mc_heatmap_condensed}  visualizes the performance of our hybrid detector across different parameter settings. The heatmap shows the total error rate (1 - accuracy), where accuracy is defined as the weighted average of true positive rate and true negative rate. Darker regions indicate lower total error rates, representing better detector performance. The axes represent different hyperparameter configurations of our detection algorithm, with optimal performance achieved in the darker regions where both false positives and false negatives are minimized.
\subsection{Visualization of Empirical Findings}
The results of the quantitative analysis would be presented through a series of visualizations for each benchmark:
\begin{itemize}
    \item Plots showing raw data points, the fitted capability curve $C(t)$, and its confidence intervals.
    \item Separate plots illustrating the time evolution of the estimated $C'(t)$, $C''(t)$, and $C'''(t)$, with their respective confidence intervals.
    \item Comparative plots showing idealized linear, exponential, and jolting growth patterns to help readers intuitively grasp the differences.
\end{itemize}
These visualizations are critical for transparently communicating the evidence for or against jolting dynamics in specific AI capabilities.

\subsection{Addressing Methodological Challenges}
\textbf{Benchmark Saturation:} The issue of benchmark saturation (where performance approaches 100\% or a practical limit) need to be explicitly addressed. Analysis would focus on periods before saturation, or on transitions to newer, harder benchmarks if data allows. The use of composite metrics that aggregate performance across multiple evolving benchmarks must also be considered \cite{ref3,ref28,ref35}.

\textbf{Sensitivity and Robustness:} To address concerns about false positives (Type-I errors), a simple sensitivity analysis has to be performed. This can involve simulating data under known growth models (e.g., exponential or logistic with varying noise levels) and observing how frequently the jolt detection methodology incorrectly identifies a jolt. This provides insight into the robustness of the findings.

\textbf{Resource Constraints in Jolt Dynamics:} While the theoretical framework includes a model for resource constraints (Equation 4), a full quantitative application is complex. The empirical analysis acknowledges this by qualitatively discussing how current resource scaling (e.g., compute trends from sources like Sevilla et al. \cite{ref29} or industry announcements \cite{ref21,ref9,ref7}) could contribute to observed acceleration, and how future resource limits could modulate jolts. A brief discussion on how a multidimensional resource model (compute, data, talent, energy) might refine the concept would be included.

\section{Case Study: AI Agent Capabilities}
\label{sec:case_study}
This section covers a case study that focuses on the capabilities of AI agents, drawing conceptual parallels with benchmarks like AgentBench \cite{ref15} and the task structures defined in METR's long-task completion research \cite{ref14,ref20}. The goal is to explore how jolting dynamics might manifest in the context of AI agents performing complex, multi-step tasks.

\subsection{Defining Agent Capability Metrics}
Measuring AI agent capability is multifaceted. For this case study, we conceptualize capability $C(t)$ as a composite metric reflecting: (1) task completion rates on a standardized set of complex tasks; (2) efficiency in task completion (e.g. steps taken, resources consumed); and (3) the complexity of tasks an agent can successfully perform. Data for such metrics would ideally be derived from standardized agent evaluation platforms or through systematic experimentation with agent architectures over time.

\subsection{Simulating Jolts in Agent Performance}
Given the challenges in obtaining long-term, consistent real-world agent performance data suitable for third-derivative analysis, our study employed simulations. We generate synthetic agent performance trajectories designed to exhibit exponential, logistic, and jolting growth patterns. These simulations allowed us to test the sensitivity of jolt detection algorithms to various forms of acceleration in agent capabilities. Factors such as the introduction of new algorithmic breakthroughs, significant increases in model scale, or improvements in agent learning strategies are modeled as potential drivers of jolts in these synthetic datasets.

We acknowledge significant challenges in obtaining the data required for full empirical validation. The correlation coefficient in Equation~\eqref{eq:composite} could be estimated through analysis of concurrent advances across AI subfields, using publication data and performance metrics from different domains. The resource coefficient R in Equation~\eqref{eq:effective} could be approximated using publicly available data on compute usage (e.g., from Epoch AI's compute trends database~\cite{epochai2023,sevilla2022}), though precise values would require industry collaboration. We propose that future work should focus on: (1) establishing partnerships with AI labs for access to detailed training logs, (2) developing proxy metrics that can be reliably measured from public benchmarks, and (3) creating synthetic datasets that capture realistic resource constraints for method validation.

\subsection{Analysis of Simulated Agent Data}
The jolt detection methodology (Savitzky-Golay smoothing, derivative estimation, hybrid detector) was applied to the simulated agent performance data. The analysis focused on identifying periods of statistically significant positive third derivatives ($C'''(t) > 0$), which would indicate a jolt. The impact of different types of simulated intervention (e.g., a step-change improvement in a core module vs. continuous improvement in learning efficiency) on the likelihood and magnitude of detected jolts was also investigated. Visualizations of these trajectories and the corresponding detected jolts are available in our repository \cite{githubJoltStudyRepo}.

\begin{figure}[htbp]
\centering
\includegraphics[width=0.8\textwidth]{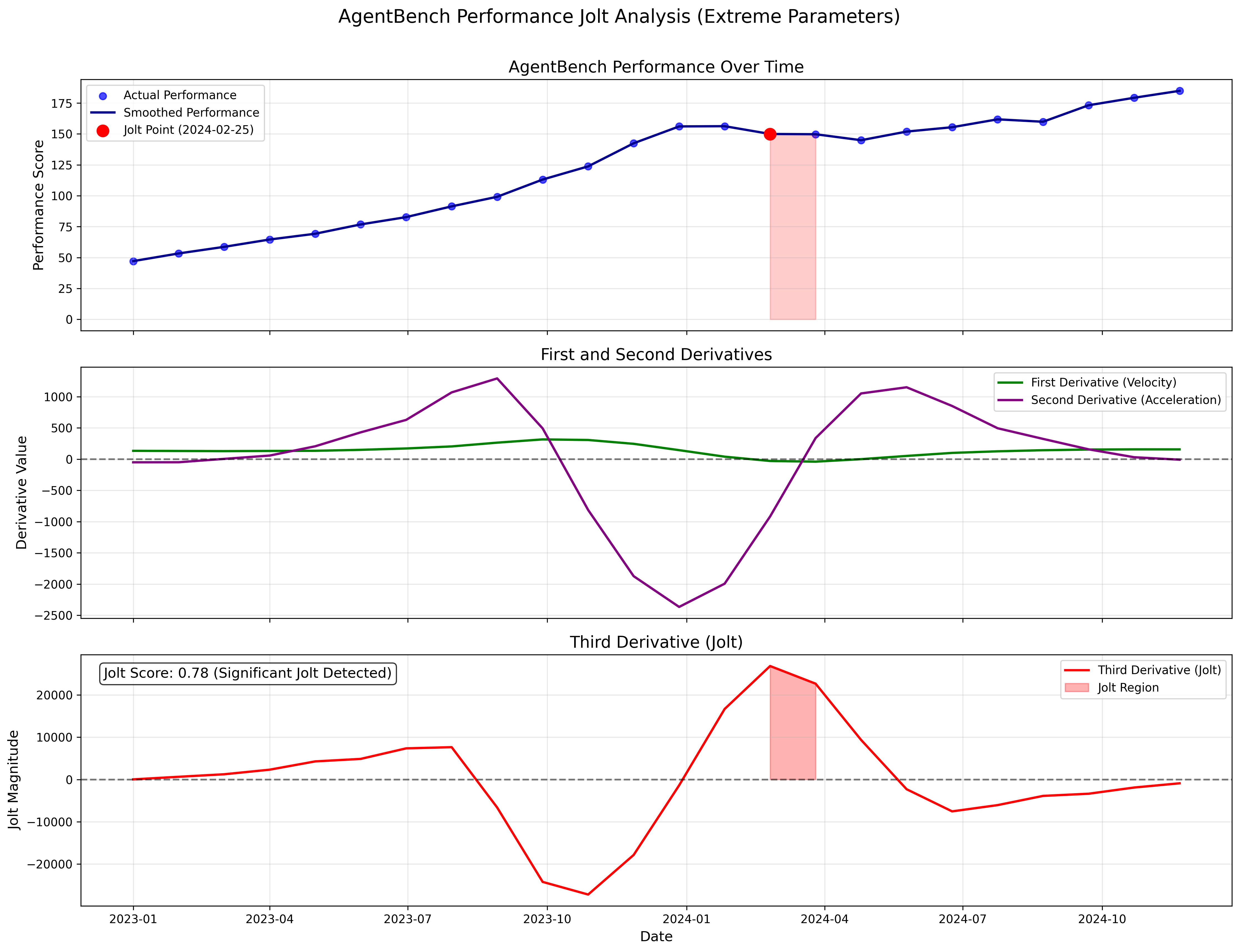}
\caption{Visualization of extreme jolt scenario in AgentBench simulation. (Further details in repository \cite{githubJoltStudyRepo}).}
\label{fig:agentbench_extreme}
\end{figure}

\subsection{Implications for Agent Development and Safety}
Real-world AI agent capabilities exhibiting jolting patterns, have significant implications. Rapid, unanticipated increases in agent capabilities pose safety challenges, if alignment and control mechanisms do not advance at a comparable pace. Furthermore, jolts in agent performance could lead to sudden changes in their economic or societal impact. This case study, through simulation, underscores the importance of developing robust monitoring and forecasting tools capable of detecting and characterizing such jolts in AI agent development.

\section{Discussion}
\label{sec:discussion}
The theoretical framework and empirical explorations presented in this paper support the Jolting Technologies Hypothesis, suggesting that AI development may be characterized by periods of superexponential growth. The ability to detect and quantify these "jolts" is critical to understanding the trajectory of AI and its implications.

\subsection{AGI Timelines and Phase Transitions}
If AI capabilities are indeed jolting, this has profound implications for the correct forecasting of AGI timelines. A sustained positive jolt implies that the rate of acceleration is itself increasing, leading to dramatically shorter doubling times for key capabilities compared to simple exponential or linear models \cite{ref1,ref14,ref20}. This necessitates a re-evaluation of existing AGI forecasts, many of which may underestimate the potential for rapid, nonlinear progress. In fact, the persistent trend of expert AGI forecasts being revised to earlier dates, as observed on platforms like Metaculus \cite{metaculusAGIforecast}, may suggest that many initial predictions were anchored in merely exponential growth assumptions, failing to fully account for the potential jolting nature of AI development. Furthermore, jolts could signify the onset of phase transitions in AI development, where qualitative shifts in capability occur rapidly and with little warning, making the emergence of AGI potentially less predictable and more abrupt than often assumed.

\subsection{AI Governance and Regulatory Preparedness}
The prospect of jolting AI development poses significant challenges for governance and regulation. Traditional governance models, which are often reactive and slow to adapt, are ill suited to manage technologies undergoing superexponential acceleration \cite{ref5}. If jolts can lead to rapid and unforeseen capability gains, then regulatory frameworks must become more agile, anticipatory, and adaptive. This could involve developing mechanisms for the continuous monitoring of AI progress, establishing flexible regulatory sandboxes, adopting sunset clauses that force the re-discussion of potentially obsolete regulations, creating rapid response teams for emerging AI risks, and investing in foresight capabilities to better anticipate potential jolts \cite{ref12}. The core challenge lies in minimizing the lag between technological advancement and effective governance responses. International cooperation will be critical to address these global challenges.

\subsection{Societal Adaptation and Economic Impacts}
Jolting AI progress accelerates the societal and economic impacts of AI. Rapid improvements in AI capabilities could lead to faster than expected labor market disruptions, requiring robust proactive measures for workforce retraining and social safety nets \cite{ref5,ref17}. The distribution of benefits and risks of jolting AI technologies also becomes a more acute concern, necessitating careful consideration of equity and access. Public discourse and education are vital to foster an informed understanding of AI's potential trajectories and to prepare society for the possibility of rapid, transformative changes \cite{ref6,ref8,ref18}.

\section{Conclusion and Future Research}
\label{sec:conclusion}
This paper has presented a theoretical framework for understanding potential superexponential growth in AI capabilities through the Jolting Technologies Hypothesis. While our Monte Carlo simulations demonstrate that jolt detection is methodologically feasible under various noise conditions, we emphasize that empirical validation with real-world data remains essential before drawing definitive conclusions about AI development trajectories. Our work provides the mathematical foundation and detection tools necessary for future empirical studies when suitable longitudinal data becomes available.

Future research should prioritize empirical investigation of the JTH using real-world AI benchmark data, refining jolt detection methodologies, and establishing statistical significance for observed jolts. Theoretical work on composite jolt models, feedback loops in AI development, and the impact of resource constraints on jolting trajectories should also continue. Crucially, interdisciplinary research involving AI scientists, ethicists, policymakers, and social scientists is essential to address the complex ethical, societal, and governance challenges posed by the prospect of jolting AI technologies.

All code, methodologies, data from our simulations, and supplementary materials referenced in this paper are available at our project repository\cite{githubJoltStudyRepo}.
\newpage
\bibliographystyle{splncs04}
\bibliography{references_v3}

\end{document}